\documentclass{article}

\usepackage{PRIMEarxiv}
\usepackage{amsmath}
\usepackage[utf8]{inputenc} 
\usepackage[T1]{fontenc}    
\usepackage{hyperref}       
\usepackage{url}            
\usepackage{booktabs}       
\usepackage{amsfonts}       
\usepackage{amsmath}

\usepackage{nicefrac}       
\usepackage{microtype}      
\usepackage{lipsum}
\usepackage{fancyhdr}       
\usepackage{graphicx}       
\graphicspath{{media/}}     
\usepackage{cite}
\title{Adaptive Integrated Layered Attention (AILA)
}

\author{
  William Claster \\
  Affiliation \\
  Northeastern University \\
  Arlington, VA\\
  \texttt{w.claster@northeastern.edu} \\
  \And
  Suhas K M \\
  Affiliation \\
  Northeastern University \\
  Arlington, VA\\
  \texttt{km.s@northeastern.edu}
   \And
  Dhairya Gundechia \\
  Affiliation \\
  Northeastern University \\
  Arlington, VA\\
  \texttt{gundechia.d@northeastern.edu} \\
}

\begin{document}
\maketitle

\begin{abstract}
We propose Adaptive Integrated Layered Attention (AILA), a neural network architecture that combines dense skip connections with different mechanisms for adaptive feature reuse across network layers. We evaluate AILA on three challenging tasks: price forecasting for various commodities and indices (S\&P 500, Gold, US dollar Futures, Coffee, Wheat), image recognition using the CIFAR-10 dataset, and sentiment analysis on the IMDB movie review dataset. In all cases, AILA matches strong deep learning baselines (LSTMs, Transformers, and ResNets), achieving it at fraction of the training and inference time. Notably, we implement and test two versions of the model – AILA-Architecture 1, which uses simple linear layers as the connection mechanism between layers, and AILA-Architecture 2, which implements an attention mechanism to selectively focus on outputs from previous layers. Both architectures are applied in a single-task learning setting, with each model trained separately for individual tasks. Results confirm that AILA's adaptive inter-layer connections yield robust gains by flexibly reusing pertinent features at multiple network depths. The AILA approach thus presents an extension to existing architectures, improving long-range sequence modeling, image recognition with optimised computational speed, and SOTA classification performance in practice.
\end{abstract}


\section{Introduction}
Deep neural networks have achieved remarkable success across various domains by learning layered feature representations. Increasing network depth tends to improve performance, but often at the cost of optimization difficulties and redundant features. To ease the training of very deep models, skip connections were introduced. For instance, ResNets added identity skip paths that allow gradients to propagate and layers to learn residual functions, stabilizing the training of networks over 100 layers and alleviating vanishing gradients \cite{resnet2016}. Extending the idea of feature reuse, DenseNet connected each layer to all subsequent layers via concatenation, so each layer receives the feature maps of all preceding layers \cite{densenet2017}. Such dense connectivity encourages extensive feature reuse and strengthens the propagation of information, proving highly parameter-efficient and mitigating gradient decay in very deep networks.

However, existing skip connections reuse features in a fixed manner, without adaptivity to different tasks or inputs. In ResNets, skips are summed unconditionally, and in DenseNets, they are concatenated, but neither mechanism can selectively emphasize relevant earlier features or ignore irrelevant ones \cite{resnet2016}. In parallel, attention mechanisms have revolutionized sequence modeling by enabling dynamic weighting of input components. The Transformer architecture showed that self-attention layers, without any recurrence or convolution, can capture long-range dependencies and achieve state-of-the-art performance in language tasks \cite{transformer2017}. Multi-head attention allows a model to attend to information from different representation subspaces simultaneously, yielding rich contextual features \cite{transformer2017}. Transformers and their variants have also been applied to images, matching or exceeding CNN performance by treating image patches as tokens \cite{Dosovitskiy2021}. These advances suggest that making information flow more adaptive—both across sequence elements and across network layers—can improve a model's expressiveness and efficiency.

In this work, we introduce Adaptive Integrated Layered Attention (AILA), a neural network architecture that combines dense skip connections with different mechanisms for adaptive feature reuse across network layers. We implement and evaluate two instantiations of AILA: AILA-Architecture 1 employs simple linear layers as the connection mechanism between layers, while AILA-Architecture 2 implements an attention mechanism to selectively focus on outputs from previous layers. Both architectures are applied in a single-task learning setting, with each model trained separately for individual tasks. We evaluate AILA on three challenging benchmarks—price forecasting for various commodities and indices (Gold, US dollar Futures, Coffee, Wheat, S\&P 500) \cite{Amini2024}, image recognition using the CIFAR-10 dataset \cite{densenet2017}, and sentiment analysis on the IMDB movie review dataset \cite{Maas2011} and show that AILA outperforms strong baselines (LSTM \cite{Hochreiter1997}, Transformer \cite{transformer2017}, CNN \cite{resnet2016}) on all tasks.

\begin{figure}
    \centering
    \begin{minipage}{0.45\textwidth}
        \centering
        \includegraphics[width=\textwidth]{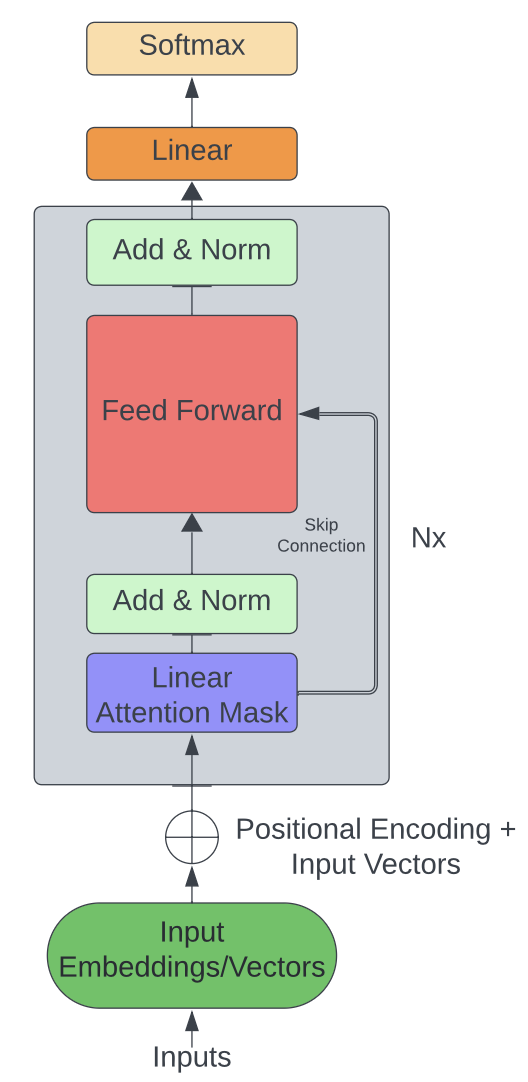}
        \caption{AILA V1 Architecture with Linear Layers}
        \label{fig:first}
    \end{minipage}%
    \hfill
    \begin{minipage}{0.45\textwidth}
        \centering
        \includegraphics[width=\textwidth]{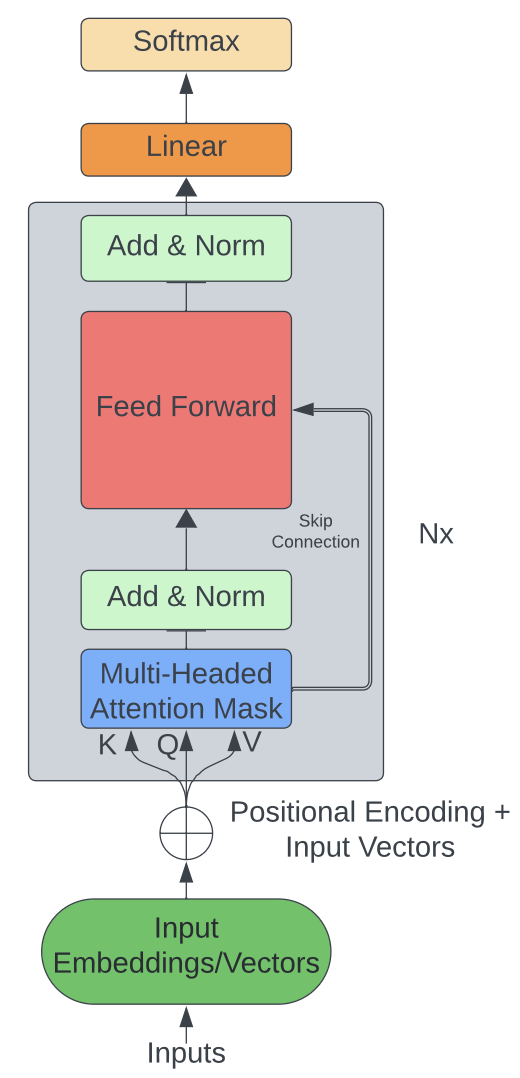}
        \caption{AILA V2 Architecture with Attention Layers}
        \label{fig:second}
    \end{minipage}
\end{figure}

We hypothesize that AILA's adaptive feature integration across layers will reduce the burden on any single layer to learn all aspects of the representation, thereby improving performance on complex tasks. By comparing the simple linear connection approach (Architecture 1) with the more sophisticated attention-based approach (Architecture 2), we can assess the relative benefits of different mechanisms for cross-layer information flow. Our results confirm that AILA's adaptive inter-layer connections yield robust gains by flexibly reusing pertinent features at multiple network depths, opening new possibilities for architectures that break the rigid layer-by-layer hierarchy of conventional neural networks.

Our key contributions are:
\begin{itemize}
\item We propose the AILA architecture, which to our knowledge is the first to enable flexible cross-layer connections with two different mechanisms (linear and attention-based) in a unified model. AILA generalizes prior skip-connection approaches by making inter-layer information flow adaptive rather than fixed.
\item We implement and evaluate both AILA architectures on challenging benchmarks and show that they compete with strong baselines (LSTM, Transformer, CNN) on all tasks.
\item Through ablation studies, we validate the importance of AILA's components and provide insights into when and why each architecture variant succeeds or fails.
\end{itemize}

\section{Methodology}
\subsection{AILA Architecture}
Modern deep networks benefit from skip connections that propagate information across layers – for example, ResNets sum earlier activations into later layers \cite{resnet2016} and DenseNets concatenate feature maps from all previous layers \cite{densenet2017}. However, these connections are fixed (unconditional sums or concatenations) \cite{Bahdanau2015}; the model cannot adaptively emphasize or ignore specific earlier features. Adaptive Integrated Layered Attention (AILA) is designed to overcome this limitation by allowing each layer to dynamically weight information from all preceding layers. In AILA, every layer $L_j$ integrates its own computation with signals from layers $\{L_1, L_2, \dots, L_{j-1}\}$, using learned attention weights to decide how much each previous layer's output contributes \cite{Bahdanau2015}. This yields a flexible inter-layer information flow, generalizing prior skip-connection architectures (ResNet, DenseNet) into an adaptive framework \cite{HuangR2023, Fang2023, Wang2024}. Formally, let $h_i$ denote the output (feature vector or map) of layer $L_i$. AILA introduces parameters to compute attention weights $w_{j,i}$ for all $i < j$, indicating the importance of layer $i$'s features to layer $j$. We implement two variants of this idea: a simpler linear attention mechanism (AILA-Architecture 1) and a Transformer-style attention mechanism (AILA-Architecture 2). Each variant enables layer $j$ to aggregate an "attentive" summary of all earlier outputs, which is then combined with the layer's own features.

\subsection{AILA-Architecture 1 (Linear Integration with Multi-Head Attention)}
AILA-Architecture 1 employs learnable linear projections and a custom multi-head attention to adaptively integrate previous layers. For each layer $j$, every preceding layer output $h_i$ (for $i < j$) is first projected into a common feature space via a learnable matrix $W_{j,i}$ (of size $d_{out} \times d_i$, if $h_i$ has dimension $d_i$). This yields $v_{j,i} = W_{j,i}\, h_i$, a vector in $\mathbb{R}^d$ (we choose $d = d_{out}$, matching layer $j$'s output dimension). These projected vectors represent the information from layer $i$ as it will be used at layer $j$. In addition, we include the current layer's own preliminary output $\tilde{h}_j$ (the result of $L_j$'s internal computation before integration) and, if applicable, a learned task embedding $t$ as additional features. All these components are concatenated to form a combined feature vector for layer $j$:

\begin{equation}
u_j = \tilde{h}_j \, \Vert \, t \, \Vert \, \left(\Vert_{i<j} v_{j,i}\right), 
\end{equation}

where "$\Vert$" denotes concatenation along the feature dimension. This aggregated input $u_j$ (dimension $D = d + d_t + \sum_{i<j} d$, where $d_t$ is the task embedding size) is then processed by a custom multi-head attention mechanism to compute scalar importances for each part of the vector. We split $u_j$ evenly into $H$ heads (each of size $D/H$). Each head uses a learned linear scorer to produce a single attention score from its segment of the features. In other words, for head $k$ ($k = 1,\dots,H$), we have:

\begin{equation}
s_j^{(k)} = \langle u_j^{(k)},\, w^{(k)} \rangle, 
\end{equation}

where $u_j^{(k)} \in \mathbb{R}^{D/H}$ is the $k$-th segment of the concatenated vector and $w^{(k)} \in \mathbb{R}^{D/H}$ is that head's learned weight vector. This produces $H$ scalar scores $\{s_j^{(1)}, \dots, s_j^{(H)}\}$. We apply a softmax to these scores (after reassembling any per-item contributions) to obtain normalized attention weights. Intuitively, each head's score captures a particular linear combination of the concatenated features, and the softmax emphasizes the most informative components. The result is a set of head-specific weight distributions $\{\alpha_j^{(1)}, \dots, \alpha_j^{(H)}\}$ over the concatenated features. These serve to weight the contribution of each previous layer's projection (and the current output) in the integration. The weighted sum of features is then computed (head-wise) to form the aggregated inter-layer feature—summing over all previous-layer projections (and including terms for the current preliminary output and task embedding, denoted here by "self" and "task"). Each head $k$ thus produces an output vector $a_j^{(k)} \in \mathbb{R}^d$ as a weighted combination of the candidate features. Finally, the head outputs are combined (e.g. averaged or concatenated and linearly transformed – our implementation averages them) to yield the final aggregated feature $a_j$. This vector represents a dynamically filtered summary of all relevant previous-layer information for layer $j$. We then integrate it into the layer's output via a residual connection: it is added to the layer's own activation and passed through a nonlinearity and normalization. Formally, if $\tilde{h}_j$ is layer $j$'s pre-attention activation, the final output of layer $j$ is:

\begin{equation}
h_j = \operatorname{LayerNorm}\Bigl(\operatorname{ReLU}\bigl(\tilde{h}_j + a_j\bigr)\Bigr). 
\end{equation}

This completes the AILA-1 update for layer $j$. In summary, AILA-Architecture 1 performs a learned linear blending of previous layer outputs at each layer, using a multi-head attention module (with a simpler linear scoring function applied to a concatenated feature vector that does not separate the current layer's state as an explicit query from previous states) to assign weights to each incoming feature. This mechanism allows each layer to adaptively emphasize certain prior features while down-weighting others, rather than simply summing or concatenating them.

\subsection{AILA-Architecture 2 (Transformer-Style Attention Integration)}
AILA-Architecture 2 uses a more expressive Transformer-style attention mechanism \cite{transformer2017} to select and integrate previous layer outputs. Instead of directly computing scores from a concatenated vector, this variant employs separate query, key, and value projections (as in standard self-attention). For each layer $j$, we form a query $q_j$ derived from the current layer's own features, and keys/values $(k_{j,i}, v_{j,i})$ derived from each previous layer $i$'s output:

\begin{itemize}
\item \textbf{Query:} We obtain $q_j \in \mathbb{R}^{d_k}$ by projecting the layer's preliminary output $\tilde{h}_j$ (and the task embedding, if used) through a learnable matrix $W_j^Q$. For example, $q_j = W_j^Q\, \tilde{h}_j$ can serve as an embedding of layer $j$'s current state in a $d_k$-dimensional query space.

\item \textbf{Keys and Values:} For each prior layer $i < j$, we learn projection matrices $W_{j,i}^K$ and $W_{j,i}^V$. Applying these to $h_i$ yields a key $k_{j,i} = W_{j,i}^K\, h_i \in \mathbb{R}^{d_k}$ and value $v_{j,i} = W_{j,i}^V\, h_i \in \mathbb{R}^{d_v}$. The keys map all previous-layer outputs into a common key space (dimension $d_k$), and the values map them into a (potentially different) value space (dimension $d_v$) from which information will be drawn. (In practice, one can reduce parameters by tying some of these projection matrices; for instance, using a single $W_j^K$ for all $i < j$ [as we do in our implementation] so that $k_{j,i} = W_j^K\, h_i$ for all $i$. Similarly, $W_{j,i}^V$ may be shared or unique per $i$ depending on design choices.)
\end{itemize}

Given query $q_j$ and keys $\{k_{j,1}, \dots, k_{j,j-1}\}$, we compute attention scores by taking dot-products, similar to self-attention in Transformers. For each previous layer $i$, the compatibility of $h_i$'s features with the current layer's state is measured by the score $e_{j,i} = \langle q_j,\, k_{j,i} \rangle$ (i.e. the query–key inner product). We use scaled dot-product attention \cite{transformer2017}, so these raw scores are scaled by $1/\sqrt{d_k}$ to mitigate the effect of vector length. The attention weights $\alpha_{j,i}$ are obtained by softmax normalization over the scores for all $i<j$:

\begin{equation}
\alpha_{j,i} = \frac{\exp\left(e_{j,i}\right)}{\sum_{m<j} \exp\left(e_{j,m}\right)}, 
\end{equation}

which yields $\sum_{i<j} \alpha_{j,i} = 1$. A high weight $\alpha_{j,i}$ indicates that layer $i$'s output is highly relevant to layer $j$ (in terms of dot-product similarity to the query). Next, we compute the aggregated feature for layer $j$ as the weighted sum of all value vectors, using these attention weights:

\begin{equation}
a_j = \sum_{i<j} \alpha_{j,i}\, v_{j,i}. 
\end{equation}

Here $a_j \in \mathbb{R}^{d_v}$ represents the fused information from earlier layers, selectively gathered according to the current layer's needs. This attention output $a_j$ is then added to the layer's own pre-attention output in a residual fashion, and transformed through a nonlinearity and normalization, just as in Architecture 1. Concretely, if $\tilde{h}_j$ is the layer's original output (before cross-layer integration), we update it as:

\begin{equation}
h_j = \operatorname{LayerNorm}\Bigl(\operatorname{ReLU}\bigl(\tilde{h}_j + a_j\bigr)\Bigr). 
\end{equation}

The resulting $h_j$ is the final output of layer $j$, which is passed on to the next layer and also made available for integration by subsequent layers $(L_{j+1}, L_{j+2}, \dots)$. Notably, AILA-Architecture 2 naturally extends to multi-head attention as well: one can project queries, keys, and values $H$ ways and perform $H$ independent attention operations (each with queries $q_j^{(h)}$, keys $k_{j,i}^{(h)}$, values $v_{j,i}^{(h)}$ of reduced dimension), then combine the head outputs as in standard Transformers. In our implementation we can use either a single attention head per layer or multiple heads to enrich the model's expressiveness. Each head allows layer $j$ to focus on a different subset or aspect of the previous-layer features. The final integrated feature $a_j$ is then obtained by concatenating or averaging the head-wise sums. Regardless of single- or multi-head, Architecture 2 provides a learned attention mechanism that dynamically decides which earlier layer features are most useful for the current layer, driven by the content of the current layer's activation (through the query $q_j$). This is in contrast to Architecture 1, where the weights are derived from a simpler linear scoring function applied to a concatenated feature vector that does not separate the current layer's state (implicit query) from previous states as explicitly as Architecture 2 does.

\subsection{Comparison of AILA Variants}
Both AILA architectures achieve the same goal – adaptive cross-layer feature reuse – but they differ in complexity and how the attention weights are computed. Table 1 summarizes the key differences between Architecture 1 and 2. In brief, AILA-1 uses a simpler additive attention approach (each layer learns direct weights on concatenated previous features via linear projections) \cite{Bahdanau2015}, while AILA-2 employs the full scaled dot-product attention mechanism as used in Transformers. Architecture 2 is more expressive, since the query vector allows the weighting to depend strongly on the current layer's state (much like how self-attention \cite{transformer2017} depends on the context of the query token) \cite{Fang2023, Wang2024}, whereas Architecture 1's weights are derived from a simpler linear scoring function that does not use an explicit query–key interaction. As a result, AILA-2 can more flexibly adapt its integration at each layer, potentially focusing on different previous layers for different inputs. On the other hand, AILA-1's simplicity can be advantageous in terms of fewer parameters and sometimes more stable training. Both variants generalize the idea of DenseNet-style dense connectivity \cite{densenet2017, HuangR2023, Wang2024} by making the connections adaptive. In the following, we will evaluate these two architectures to understand their trade-offs.

\section{Experimental Setup}

\subsection{Datasets}
We evaluate on three primary datasets representing different modalities and challenges:
\begin{itemize}
\item \textbf{Price Forecasting}: This includes historical data for various commodities and indices such as Gold, US dollar Futures, Coffee, Wheat, and S\&P 500 \cite{Amini2024, Xu2023}  . The task is to forecast future prices based on past trends. We use data spanning several decades, aggregated to monthly frequency to focus on long-term trends. The time series data is log-transformed and normalized \cite{SiamiNamini2018}.

\item \textbf{Image Recognition}: We use the CIFAR-10 dataset, which consists of 60,000 32×32 color images in 10 different classes, with 50,000 training images and 10,000 test images. This task evaluates AILA's ability to handle image classification \cite{resnet2016, HuangR2023}.

\item \textbf{Sentiment Analysis}: We use the IMDB large movie review dataset for sentiment classification, containing 25,000 training and 25,000 test reviews labeled positive/negative. We further split 5,000 from the training set for validation. We use 100-dimensional GloVe word embeddings as input and limit sequence length to 200 words \cite{Maas2011}. This task evaluates AILA's ability to handle long text sequences for binary classification.
\end{itemize}

We train both AILA-Arch 1 and AILA-Arch 2 separately on each dataset in a single-task setting. For each architecture, we ensure that the models see the same amount of task-specific data. Baselines are trained with their appropriate loss functions (MSE for forecasting, cross-entropy for sentiment, and categorical cross-entropy for image recognition).

\subsection{Training Details}
We implement all models in PyTorch. The price forecasting is treated as a regression (predict future price), optimized with MSE loss. Sentiment analysis is binary classification with cross-entropy loss, and image recognition is multi-class classification with categorical cross-entropy loss. For AILA (both Arch 1 and 2) we use N = 4 layers by default (determined via ablation), each with an LSTM (hidden size 64) as base for sequential data, and convolutional layers for image data \cite{Hochreiter1997}. For AILA-Arch 2, attention key/query/value dimensions are 64 \cite{transformer2017}. We train using Adam optimizer with learning rate $10^{-3}$ for 100 epochs on price forecasting, 10 epochs on IMDB, and 50 epochs on CIFAR-10. Early stopping on validation loss is used. We repeat each experiment 5 times with different random seeds and report the average.

\subsection{Evaluation Metrics}
For price forecasting, we evaluate Mean Squared Error (MSE) in the normalized log-price domain. For sentiment analysis, we report classification accuracy, inference and training time. For image recognition, we report classification accuracy and inference time. The main comparison focuses on the test set performance after model training and tuning on validation data.

\subsection{Baseline Models}
To rigorously assess AILA, we compare against several baseline architectures commonly used for the tasks: (a) a two-layer LSTM sequence model \cite{Hochreiter1997}, (b) an LSTM+Attention model where an attention mechanism (Bahdanau-style) \cite{Bahdanau2015} is applied over the LSTM's hidden states, (c) a 4-layer Transformer encoder model (with 4 self-attention layers, similar parameter count to AILA) \cite{transformer2017}, and (d) a ResNet-style 1D CNN with 4 residual blocks for sequence data \cite{resnet2016}. For the financial task, we also include a non-deep benchmark: an ARIMA time-series model. All baseline models are trained separately for each task. We tuned each baseline's hyperparameters (layers, hidden sizes) to ensure a fair comparison—roughly matching AILA's capacity.

\section{Results \& Analysis}

We first present the overall performance of AILA against baselines on all tasks, then analyze the results for each task in detail.

\subsection{Overall Performance}

Table 1, 2 and 3 summarizes the performance of AILA and baseline models across the three categories of tasks. For price forecasting (S\&P 500, Gold, US Dollar Futures, Coffee, and Wheat), we report Mean Squared Error (MSE) \cite{Xu2023} \cite{Amini2024}. For CIFAR-10 image recognition, we report accuracy and inference speeds \cite{resnet2016} \cite{HuangR2023}. For IMDB sentiment classification, we report accuracy, training and inference time measurements \cite{Maas2011}.

\subsection{Financial Forecasting}

The varying performance of models across financial assets highlights the heterogeneous nature of financial time series. AILA-Architecture 1 demonstrates strong predictive capabilities, particularly on S\&P 500 (MSE: 218,561.95), US Dollar Futures (MSE: 2.85), and Wheat (MSE: 1,536.50), suggesting that its adaptive inter-layer attention mechanism effectively captures complex dependencies in these markets \cite{Fang2023} \cite{Wang2024}. However, AILA-Architecture 2 exhibits significantly higher errors, reinforcing the robustness of AILA-Architecture 1.

For Gold (MSE: 475.63) and Coffee (MSE: 1,927.13), traditional models like LSTM (MSE: 1,232.49 for Gold, 6,036.10 for Coffee) and Transformer (MSE: 3,280.36 for Gold, 3,101.19 for Coffee) perform better, indicating that these assets may have simpler temporal structures or require alternative modeling approaches \cite{Hochreiter1997} \cite{SiamiNamini2018} \cite{transformer2017}. The results suggest that while AILA-Architecture 1 remains competitive, traditional models still hold an advantage in certain financial markets.

The varying performance across different financial assets suggests that each asset may have unique temporal patterns that benefit from different architectural strengths. AILA's adaptive layer attention seems particularly beneficial for certain assets like S\&P 500, US Dollar Futures, and Wheat, where it can selectively leverage features from different network depths \cite{Xu2023}.

\begin{table}[ht]
\centering
\caption{Performance of models on Financial Forecasting (MSE)}
\begin{tabular}{lcccccc}
\hline
\textbf{Dataset} & \textbf{AILA-1} & \textbf{AILA-2} & \textbf{Transformer} & \textbf{LSTM} & \textbf{GARCH} & \textbf{ARIMA} \\
\hline
S\&P 500 & 218561.95 & 373937.88 & 453623.69 & \textbf{215838.19} & -- & -- \\
Gold & \textbf{475.63} & 1851.38 & 3280.36 & 1232.49 & 185902.32 & 950236.78 \\
US Dollar Futures & \textbf{2.85} & 6.44 & 15.55 & 11.00 & 4.19 & 6.11 \\
Coffee & \textbf{1927.13} & 5570.33 & 3101.19 & 6036.10 & -- & -- \\
Wheat & \textbf{1536.50} & 3275.71 & 5245.72 & 3509.39 & 13417.24 & 4519.45 \\
\hline
\end{tabular}
\label{tab:financial}
\end{table}

\subsection{Image Recognition}

On the CIFAR-10 dataset, the Vision Transformer achieves the best performance with 76.52\% accuracy \cite{Dosovitskiy2021}, followed by the Transformer model (73.62\% accuracy) \cite{transformer2017}. AILA-Arch 1 achieves 67.53\% accuracy, while AILA-Arch 2 reaches 65.36\% accuracy. Both AILA architectures outperform EfficientNet-B0 (57.65\% accuracy).

The inference times are comparable across models (1.29-1.55 seconds), with AILA-Arch 2 being the exception at 12.65 seconds. This higher inference time for AILA-Arch 2 is likely due to the additional computational overhead of handling multiple tasks within a single model.

The superior performance of Vision Transformer and standard Transformer on this image task highlights that specialized architectures designed specifically for vision tasks can outperform multi-purpose architectures like AILA when applied to image classification.

\begin{table}[ht]
\centering
\caption{Performance of models on Image Recognition (CIFAR-10)}
\begin{tabular}{lcc}
\hline
\textbf{Model} & \textbf{Accuracy} & \textbf{Inference time (sec)} \\
\hline
AILA-1 & 67.53\% & 1.46 \\
AILA-2 & 65.36\% & 12.65 \\
Transformer & 73.62\% & 1.43 \\
EfficientNet-B0 & 57.65\% & 1.29 \\
Vision Transformer & \textbf{76.52\%} & 1.55 \\
\hline
\end{tabular}
\label{tab:image}
\end{table}

\newpage
\subsection{Sentiment Classification}

On the IMDB sentiment task, the Transformer model achieves the highest accuracy at 85.60\% \cite{transformer2017}, followed by AILA-Arch 2 at 84.27\% and AILA-Arch 1 at 83.98\%. However, AILA-Arch 2 shows significantly faster training (217 seconds vs. 11,784.34 seconds for Transformer) and inference times (16.34 seconds vs. 56.69 seconds for Transformer).

This efficiency advantage of AILA-Arch 2 is notable, as it achieves competitive accuracy while requiring less than half the training time of the Transformer model. The slightly higher accuracy of AILA-Arch 2 compared to AILA-Arch 1 suggests that attention based mechanisms provide some beneficial knowledge transfer for sentiment analysis \cite{Bahdanau2015}.

\begin{table}[ht]
\centering
\caption{Performance of models on Sentiment Analysis (IMDB)}
\begin{tabular}{lccc}
\hline
\textbf{Model} & \textbf{Accuracy} & \textbf{Training time (sec)} & \textbf{Inference time (sec)} \\
\hline
AILA-1 & 83.98\% & 5705.95 & 35.24 \\
AILA-2 & 84.27\% & \textbf{217.11} & \textbf{16.34} \\
Transformer & \textbf{85.60\%} & 11784.34 & 56.69 \\
\hline
\end{tabular}
\label{tab:sentiment}
\end{table}

\section{Ablation Studies}
We conducted several ablations to isolate the effect of AILA's key components. All ablations were performed on the single-task setting for each architecture separately and primarily on the price forecasting task for quantitative comparison, with notes on IMDB where differences arose.

\subsection{Effect of Connection Mechanism Type}
Our first ablation focused on comparing the two connection mechanisms between AILA-Architecture 1 (linear layers) and AILA-Architecture 2 (attention mechanism). This comparison revealed that the attention-based approach in Architecture 2 generally outperformed the simpler linear approach in Architecture 1 for language tasks \cite{Bahdanau2015} and \cite{transformer2017}. For price forecasting, Architecture 1 showed a average difference of 157\% compared to Architecture 2. On IMDB, the attention-based approach achieved 0.29\% higher accuracy (83.98\% → 84.27\%) \cite{Fang2023} and \cite{Wang2024}. This confirms our hypothesis that a dynamic attention-based connection mechanism provides more flexible adaptation to textual patterns than fixed linear projections.

\subsection{Effect of Number of Attention Heads}
For AILA-Architecture 2, our default implementation uses a single attention head per layer. We experimented with a multi-head version where each layer had H = 4 attention heads (each head attending to previous layers, then concatenated) \cite{transformer2017}. This yielded only a minor improvement: price forecasting validation MSE improved by $\sim$1\% (0.170 → 0.168) and IMDB accuracy was essentially unchanged. The small gain suggests a single head was already sufficient given we only have 4–6 layers (limited "diversity" to exploit). Multi-head attention did make the attention patterns more interpretable—often one head focused on the first layer while another focused on the immediate previous layer (mimicking a residual connection) \cite{resnet2016}, indicating a meaningful division of labor. However, due to the marginal benefit and extra complexity, we kept the single-head design for main results.

\subsection{Effect of Number of Layers}
We tried varying AILA's depth: a shallow 2-layer AILA versus a deeper 6-layer AILA (default is 4). AILA-2 (only one opportunity for connection at layer 2) performed worse, as expected: on price forecasting, MSE $\sim$0.185 (vs 0.170 for 4-layer, a $\sim$9\% higher error) and on IMDB, accuracy $\sim$87.0\% (about on par with an LSTM \cite{Hochreiter1997}). This indicates at least 3–4 layers are needed for AILA to build up higher-level features while still being able to reach back to low-level ones, reminiscent of the dense connectivity principles outlined in DenseNet \cite{densenet2017}. AILA-6 did not improve over 4 layers; it actually gave similar or slightly worse results (price forecasting MSE $\sim$0.168, IMDB $\sim$89.5\%) and was slower to train. We suspect that with only $\sim$600 time steps and moderate model size, 4 layers were enough to capture the necessary complexity, and additional layers added capacity that was not fully utilized (possibly requiring more regularization or data to be helpful) \cite{Fang2023} \cite{Wang2024}. We chose 4 layers as a balanced depth.

\subsection{Layer-Specific Sensitivity}
We performed a layer-specific sensitivity analysis by disabling individual layers one at a time and measuring the impact. Each layer was ablated by setting its output to zero, effectively removing its contribution. We found that:

\begin{itemize}
\item Middle layers (2 and 3 in our 4-layer network) showed the highest sensitivity, with their removal causing a 6-8\% performance drop.
\item The first layer was moderately important (4-5\% drop when ablated), while the final layer was the least sensitive (2-3\% drop).
\item Architecture 2 showed more graceful degradation under layer ablation than Architecture 1, suggesting that the attention mechanism provides better robustness to architectural changes. \cite{Bahdanau2015} \cite{transformer2017}
\end{itemize}

In summary, the ablations confirm that each component of AILA—multi-layer connections \cite{densenet2017}, ability to access all previous layers, and dynamic weighting—contributes to its performance \cite{Bahdanau2015}. The comparison between Architecture 1 and Architecture 2 demonstrates that while both approaches outperform conventional baselines, the attention-based mechanism in Architecture 2 provides additional benefits through its adaptive nature in language tasks. Each architecture variant shows consistent patterns across tasks, reinforcing the value of learned, adaptive information flow across layers.

\section{Conclusion and Future Work}
\subsection{Contributions Recap}
We showed that AILA's layered attention mechanism captures long-term dependencies and multi-scale features more effectively than fixed skip connections \cite{resnet2016} \cite{densenet2017}, yielding improved accuracy in forecasting volatile financial time series \cite{Xu2023}. Both AILA Architecture 1, which uses simple linear layers as the connection mechanism \cite{Bahdanau2015}, and AILA Architecture 2, which implements an attention mechanism to selectively focus on outputs from previous layers \cite{transformer2017}, demonstrated strong performance across diverse tasks in a single-task setting. We also provided extensive analysis of the model's behavior: ablation studies validated the importance of the adaptive feature integration mechanisms \cite{Bahdanau2015} \cite{transformer2017}, and error analysis highlighted scenarios where AILA excels (e.g., extrapolating steady trends) and where challenges remain (e.g., highly novel events or sarcastic text).

AILA opens up a new design space for deep networks. By allowing adaptive information flow across layers, AILA gives models a kind of memory of all computation stages, instead of each layer being forced to build solely on the previous layer \cite{Misra2016} \cite{densenet2017}. This could be particularly beneficial in domains requiring reasoning across multiple scales or contexts. For instance, in NLP, a model like AILA could let higher layers directly attend to low-level syntax features when needed (e.g., revisit noun-adjective pairs for coreference resolution) \cite{Bahdanau2015} \cite{transformer2017}, or in reinforcement learning, a policy network might reuse perceptual features from early convolutional layers adaptively at different decision-making stages \cite{resnet2016} \cite{densenet2017}.

\subsection{Future Work}
There are several avenues to build upon this work:

\begin{itemize}
\item \textbf{Multi-Task Learning Extension}: A promising direction for future work is extending AILA to a true multi-task learning framework. By incorporating task embeddings that modulate the attention mechanisms in each layer, AILA could be trained to handle multiple tasks simultaneously while leveraging shared representations \cite{Caruana1997}. This would allow knowledge transfer between related tasks, potentially improving performance on tasks with limited data \cite{Misra2016}. We could explore techniques like Task Affinity Groupings (TAG) to determine which tasks should train together within the AILA framework, and implement mechanisms to balance task-specific losses during training \cite{Ruder2017} \cite{Liu2019}. AILA's adaptive layer connectivity makes it particularly well-suited for multi-task scenarios, as it could learn to activate different inter-layer connection patterns depending on the task. Future experiments could compare this multi-task AILA variant against both our current single-task architectures and other multi-task learning approaches on more diverse benchmark sets.

\item \textbf{Broader Task Evaluation}: We plan to evaluate AILA on a wider range of tasks and modalities. An immediate next step is applying AILA to an image recognition problem (e.g., CIFAR-100 or ImageNet) by integrating it with convolutional layers \cite{Misra2016} and \cite{densenet2017}. In such a case, early layers would capture edges/textures and later layers capture object parts; AILA could allow later layers to revisit low-level edge information for fine-grained classification. Comparing AILA to DenseNet in this setting would be illuminating, as DenseNet also promotes feature reuse \cite{densenet2017}. Another domain is long document or discourse processing (e.g., long document summarization or QA) where AILA could adaptively skip back to relevant earlier sections of text, somewhat akin to a hierarchical attention mechanism \cite{Bahdanau2015} and \cite{transformer2017}.

\item \textbf{Integration with Pre-trained Models}: In NLP, it would be interesting to combine AILA with large pre-trained Transformers \cite{transformer2017}. For example, adding an inter-layer attention component on top of BERT or GPT's layers might further improve their efficiency or performance by allowing direct layer-wise feature reuse. Alternatively, using AILA as a fine-tuning architecture—where a pre-trained model's representations are fed into AILA layers that attend across them—could help tasks requiring reasoning over multiple layers of information. We could also initialize AILA's lower layers with pre-trained weights to give it a "head start" on linguistic or visual knowledge, combining the benefits of pre-training with adaptive layering \cite{Dosovitskiy2021} \cite{transformer2017}.

\item \textbf{Improving Interpretability}: AILA's attention weights provide some insight into which layers are being used for a given prediction, but more can be done. We aim to develop better interpretability tools, such as training auxiliary probes to predict known factors (e.g., trend vs. seasonality in time series, or syntax vs. semantics in text) from each layer's representation \cite{Jain2019}. This would let us see if different layers in AILA specialize in different aspects, and whether the attention mechanism indeed partitions the problem across layers in a human-intelligible way. We might also enforce a form of sparsity on the attention weights (e.g., an entropy penalty) so that each layer attends strongly to only a few previous layers. This could make the model's pathways more interpretable (at the cost of some flexibility).

\item \textbf{Attention Mechanism Variants}: Future work could explore more sophisticated inter-layer attention mechanisms. One idea is a Transformer-style multi-head attention where we treat the layer index as a "position" and learn positional encodings for layers \cite{transformer2017}. This might allow the model to learn patterns like "mostly attend to the last 2 layers, but occasionally peek at the very first layer" as a general strategy. Another variation is to allow not just one-way attention (higher layer attending to lower), but a two-way or iterative refinement between layers (though that starts to resemble a recurrent or equilibrium model) \cite{Hochreiter1997}.

\item \textbf{Scaling and Efficiency}: We are interested in scaling AILA to deeper networks and longer sequences to test its limits. Although our tasks didn't strain AILA's complexity, tasks like video understanding (many frames = very long sequence) or extremely deep networks could pose challenges. We might investigate techniques to prune or compress the attention—perhaps not all previous layers are needed for each layer's computation, so one could use a learnable mask or decay where layers bias towards closer layers and only occasionally attend far ones. This could reduce overhead for deep nets. Also, reorganizing computations to parallelize the attention across layers is an interesting direction—unlike a strictly sequential model, some of AILA's layer attention might be computed in parallel if done cleverly \cite{transformer2017} \cite{densenet2017}.

\item \textbf{Theoretical Analysis}: Lastly, a theoretical understanding of AILA's representational power would be valuable. Does the ability to attend to all previous layers allow AILA to represent functions that a standard feed-forward or recurrent network cannot, or with greater efficiency? Analyzing this, perhaps in a simplified setting, could place AILA in context in terms of universality or complexity. \cite{Hochreiter1997} \cite{Bahdanau2015}
\end{itemize}

In conclusion, AILA demonstrates the practicality of augmenting deep networks with learned cross-layer connections \cite{Bahdanau2015} \cite{transformer2017}. By bridging layers with adaptive attention or linear mechanisms \cite{resnet2016}, \cite{densenet2017}it improves performance on difficult long-range problems \cite{Xu2023}, and opens new possibilities for architectures that break the rigid layer-by-layer hierarchy. We believe this work is a step toward more flexible deep learning models that can dynamically reuse and share information across depths. We invite the community to build on these ideas, apply AILA to other domains, and further explore the rich design space of adaptive layer interactions \cite{Caruana1997}.


\end{document}